\pdfoutput=1
\documentclass[11pt]{article}

\usepackage[final]{coling}

\usepackage{times}
\usepackage{latexsym}
\usepackage{float}
\usepackage{booktabs}
\usepackage[T1]{fontenc}
\usepackage[utf8]{inputenc}

\usepackage{microtype}

\usepackage{inconsolata}

\usepackage{graphicx}

\usepackage{array,booktabs}

\newcolumntype {+}{ >{\global\let\currentrowstyle\relax}}
\newcolumntype {^}{ >{\currentrowstyle }}

\title{Funzac at CoMeDi Shared Task: Modeling Annotator Disagreement from Word-In-Context Perspectives}

\author{\textbf{Olufunke O. Sarumi\textsuperscript{1}},
    \textbf{Charles Welch\textsuperscript{2}},
    \textbf{Lucie Flek\textsuperscript{3}},
    \textbf{Jörg Schlötterer\textsuperscript{1,4}}
    \\ 
    \textsuperscript{1}University of Marburg, \textsuperscript{2}McMaster University, \textsuperscript{3}University of Bonn, \textsuperscript{4}University of Mannheim\\
    \small\{sarumio,joerg.schloetterer\}@uni-marburg.de\textsuperscript{1}, cwelch@mcmaster.ca\textsuperscript{2}, flek@bit.uni-bonn.de\textsuperscript{3}
}

\begin{document}
\maketitle
\begin{abstract}
In this work, we evaluate annotator disagreement in Word-in-Context (WiC) tasks exploring the relationship between contextual meaning and disagreement as part of the CoMeDi shared task competition. 
While prior studies have modeled disagreement by analyzing annotator attributes with single-sentence inputs, this shared task incorporates WiC to bridge the gap between sentence-level semantic representation and annotator judgment variability. 
We describe three different methods that we developed for the shared task, including a feature enrichment approach that combines concatenation, element-wise differences, products, and cosine similarity, Euclidean and Manhattan distances to extend contextual embedding representations, a transformation by Adapter blocks to obtain task-specific representations of contextual embeddings, and classifiers of varying complexities, including ensembles.
The comparison of our methods demonstrates improved performance for methods that include enriched and task-specfic features.
While the performance of our method falls short in comparison to the best system in subtask 1 (OGWiC), it is competitive to the official evaluation results in subtask 2 (DisWiC).

\end{abstract}

\section{Introduction}
Disagreement in annotation tasks has been widely studied, with various methods proposed to address it \cite{leonardelli-etal-2023-semeval}. One of the most common approaches is majority voting \cite{nguyen-etal-2017-aggregating}, where the most frequently chosen annotation is treated as the correct label. 
Recent research explores alternatives to this traditional majority voting paradigm, modeling individual annotators and their labels to predict perspectives, aiming to account for individual differences in judgment \cite{plepi-etal-2022-unifying, 10.1162/tacl_a_00449, oluyemi-etal-2024-corpus} and exploring the use of demographic information to cluster annotators, using these clusters to model disagreement \cite{deng-etal-2023-annotate}. However, fewer authors considered the role of contextual information in pairwise sentences, which can shed light on the root causes of disagreement \cite{pilehvar-camacho-collados-2019-wic, armendariz-etal-2020-semeval}. Understanding these causes may reveal ambiguities in data and help to gain insights into why annotators diverge in their judgments.

While not explicitly posed as such, we view the CoMeDi shared task~\cite{schlechtweg2025comedi} in light of these recent trends, offering potential avenues for a better understanding of contextual ambiguities and their consequences on annotator disagreement.
This shared task  involves modeling disagreement in word sense annotation for the Word-in-Context (WiC) task, where annotators provide judgments on the relatedness of two word uses in a sentence pair, rated on an ordinal scale from 1 (homonymy) to 4 (identity). It includes two subtasks: Median Judgment Classification, which predicts the median of annotator ratings as an ordinal classification task evaluated with Krippendorff’s $\alpha$, and Mean Disagreement Ranking, which quantifies the magnitude of disagreement between annotators by ranking instances based on pairwise absolute differences evaluated with Spearman's \( \rho \). 
From the methods we developed, the inclusion of task-specific representations obtained by transformations of contextual embeddings via Adapter blocks outperformed our other methods in predicting the median in the OGWiC task, In the DisWiC task, the best performance among our approaches alternated between this method and an ensemble of XGBoost and CatBoost on enriched feature combinations of contextual embeddings. 

We made submissions to the shared task at the post evaluation phase and make our implementation publicly available.\footnote{\url{https://github.com/funzac/comedi}}

\section{Shared Task}
The shared task is subdivided into two subtasks, Median Judgment Classification with Ordinal Word-in-Context Judgments (OGWiC) and  Mean Disagreement Ranking with Ordinal Word-in-Context Judgments (DisWiC). In both tasks, a training instance consists of (i) a pair of two contexts (each context is a sentence or paragraph), (ii) a target word (lemma) that appears in both contexts, (iii) ordinal ratings by multiple annotators of how related the meanings of the lemma are in the two contexts on a scale from 1 (completely unrelated) to 4 (identical). Each instance contains additional information on the language of contexts, lemmas, and indices of the target word.
The two tasks differ in their prediction targets:
\begin{description}
    \item[OGWiC] Predict the median rating. Predictions are evaluated by the ordinal version of Krippendorf's $\alpha$ against the ground truth median ratings.
    \item[DisWiC] Predict the mean disagreement, i.e., the mean of average pairwise differences in relatedness ratings and rank by magnitude of disagreement. Predictions are evaluated by Spearman's $\rho$ against ground truth disagreement ranking.
\end{description}

\section{System Description}
Following the setup of the baseline method provided by the task organizers, our system builds upon contextual embeddings of the lemma in both contexts, obtained from the XLM-RoBERTa (XLM-R\footnote{\url{https://huggingface.co/FacebookAI/xlm-roberta-base}}) transformer model~\citep{conneau2020unsupervisedcrosslingualrepresentationlearning}.
We investigated three methods (XLM-R, XLMR + Ensemble, XLM-R + Adapter), featuring different classifiers in the ordinal classification task OGWiC and different regressors in the DisWiC task.
We additionally enriched the input to XLMR + Ensemble and XLM-R + Adapter by pairwise comparisons of the contextual embeddings, such as element-wise difference.
The XLM-R + Adapter method further includes the transformation of the contextual embeddings in the input to a task-specific representation. 
%which serve as inputs to the three models developed for the two subtasks in the CoMeDi shared task: OGWiC as an ordinal classification task and DisWiC as a regression task. In the (XLM-R + ensemble) and (XLM-R + Adapter) models, we enriched the embeddings with additional features building on \cite{reimers2019sentencebertsentenceembeddingsusing} for pairwise comparisons—such as concatenation, element-wise differences, product, cosine similarity, and distance metrics for a broader embedding representation.

\subsection{CoMeDi Baselines}
\label{ssec:baselines}
The baseline methods provided by the task organizers start from contextual embeddings $e_1$ and $e_2$ of the lemma in context 1 and context 2 respectively. These contextual embeddings are obtained from a pre-trained XLM-RoBERTa model. Specifically, $e_1$ and $e_2$ are the mean of the last hidden states of the hidden states corresponding to the subword tokens of the lemma in each respective context.

In the DisWiC task, the contextual embeddings are concatenated to obtain an input representation $f=[e_1|e_2]$ (where $|$ denotes concatenation) for a Linear Regression model. The dependent variable in the linear regression is the average disagreement of annotators.

In the OGWiC task, the organizers first calculate the cosine similarity between $e_1$ and $e_2$ and place them into four bins, corresponding to the median judgement values. The bin boundaries are directly optimized with respect to the target measure of the task, Krippendorf's $\alpha$.

\subsection{XLM-R}
\label{ssec:xlmr}
Our XLM-R method uses the concatenation of contextual embeddings $f=[e_1|e_2]$ as input in both, the OGWiC classification and the DisWiC regression task.

\begin{figure}
    \centering
    \includegraphics[width=\linewidth, trim={0 0 0 0.7cm},clip]{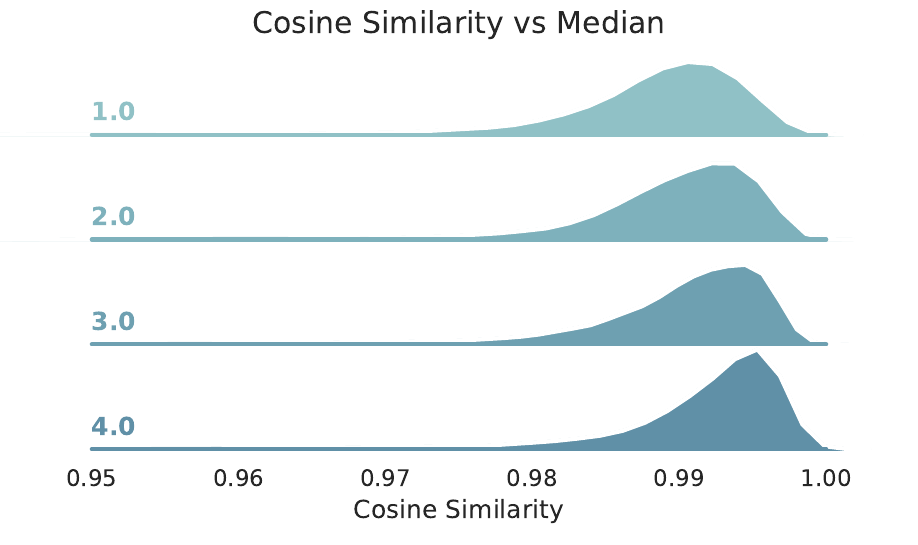}
    \caption{Densities of cosine similarity (x-axis) of context embeddings $e_1$ and $e_2$ vs median similarity rating (y-axis). Note that the x-axis does not start at 0.}
    \label{fig:cosine-bins}
\end{figure}

Analyzing the cosine similarities between pairs of contextual embeddings ($e_1$ and $e_2$) in the OGWiC task, we discovered that these are hardly separable into distinct bins (see Figure~\ref{fig:cosine-bins}).
Therefore, we decided to cast the task as multi-class classification, aiming to predict the median similarity judgement per instance. On the concatenation of contextual embeddings $f=[e_1|e_2]$, we train a simple linear classification head with dropout.

This method for the DisWiC task is almost identical to the baseline, only adding dropout to the linear regression head.

\subsection{Feature Enrichment}
\label{ssec:enrichment}
Inspired by \citet{reimers2019sentencebertsentenceembeddingsusing}, we enrich the original input $f=[e_1|e_2]$, i.e., the concatenation of contextual embeddings, by pairwise comparisons and similarity measures of the two embeddings. Specifically, we extend $f$ to $f_e = [e_1|e_2|e_1-e_2|e_1*e_2|C|E|M]$ where "$-$" and "$*$" indicate element-wise difference and multiplication, and $C$, $E$, and $M$ indicate cosine similarity, Euclidean and Manhattan distance. We use this extended feature representation $f_e$ as input in both, XLM-R + Adapter and XLM-R + ensemble for both tasks (OGWiC and DisWiC).

\subsection{XLM-R + Adapter}
\label{ssec:xlmr+adapter}
In this method, we first transform the original contextual embeddings $e1$ and $e_2$ in the input $f_e = [e_1|e_2|e_1-e_2|e_1*e_2|C|E|M]$ (cf. section~\ref{ssec:enrichment}) to task-specific representations $e_1'$ and $e_2'$, followed by a classification/regression network on the adapted representations $f_a = [e_1'|e_2'|e_1|e_2|e_1-e_2|e_1*e_2|C|E|M]$. For the transformation, we use the architecture of adapter blocks~\citep{houlsby2019parameterefficienttransferlearningnlp}, which is a bottleneck architecture with down-projection, GELU activation, dropout for regularization, up-projection, and a residual connection. We use a separate adapter block for each transformation $e_1 \rightarrow e_1'$ and $e_2 \rightarrow e_2'$.

The classification/regression network consists of two hidden layers of size 512 and 256 with GELU activation, each preceded by layer normalization and followed by dropout, and a final linear classification (OGWiC) or regression (DisWic) head.

The adapter blocks are jointly trained with the classification/regression network, turning the contextual embeddings into a task-specific representation: While the contextual embeddings are obtained from a frozen XLM-RoBERTa model optimized for language modeling, their transformation is optimized for the classification/regression task.

\subsection{XLM-R + Ensemble}
We train the two ensemble methods, CatBoost \cite{prokhorenkova2019catboostunbiasedboostingcategorical} and XGBoost \cite{Chen_2016}, independently on the enriched input $f_e = [e_1|e_2|e_1-e_2|e_1*e_2|C|E|M]$ (cf. section~\ref{ssec:enrichment}).
We then combine their predictions, effectively forming an ensemble of ensembles. In the OGWiC task, we weigh the predictions of the CatBoost and XGBoost classifiers with 0.4 and 0.3 in the combined prediction (linear combination). In the DisWiC task we weigh the CatBoost and XGBoost regressors with 0.4 and 0.6 (weighted average).

\subsection{Hyper-parameters}
We train all our networks (including adapter blocks) for 10 epochs with a learing rate of 1e-4, AdamW optimizer, batch size of 32 and dropout rate of 0.2.

We train both ensemble models (XGBoost and CatBoost) with a learning rate of 0.05, a maximum depth of 6, and 500 iterations/estimators. Additionally, we set the column sub-sampling rate in XGBoost to 0.8. 

We keep all other hyper-parameters at the default values provided by their respective libraries.

\section{Dataset}
Separate datasets were provided for OGWiC and DisWiC task. However, the uses of a word, i.e., a lemma in one particular context are identical for both tasks. That is, both tasks have the same set of available contexts and lemmas. Yet, the instances per task differ in the extent that they make use of the combinatorial options to combine different contexts for the same lemma and do not necessarily make use of all combinatorial options (probably due to the unavailability of ratings). From what we observed, instances in the OGWiC task are a subset of the instances in DisWiC, discarding instances where no meaningful median of the ratings can be obtained.
For both tasks (OGWiC and DisWiC), the datasets were divided into pre-defined train, dev and test splits. The OGWiC task data includes 47.8K training, 8.3K dev and 15.3K test instances in different languages from prior work, specifically Chinese~\cite{Chen2023chiwug}, German~\cite{Schlechtweg2024dwugs}, Russian~\cite{rushifteval2021,Aksenova2022rudsi}, English~\cite{Schlechtwegetal18}, Swedish~\cite{Schlechtweg2024dwugs}, Spanish~\cite{Zamora2022lscd}, and Norwegian~\cite{kutuzov2022nordiachange}. 
The DisWiC task data includes 82.2K training, 13.1K dev and 26.7K test instances from the same languages.
Table~\ref{tab:dataset} details the training set statistics per language.
\begin{table*}[tbh]
\small
\centering
\begin{tabular}{lcccccccc}
\toprule
 & AVG & ZH & DE & EN & NO & RU & ES & SV \\
 \midrule
\multicolumn{9}{c}{\textbf{Available Set of Contexts and Lemmas}} \\
 \midrule
Unique Contexts & 7,844 & 1,119 & 12,141 & 6,565 & 1,222 & 24,848 & 2,757 & 6,256\\
Unique lemmas & 74 & 28 & 117 & 31 & 56 & 189 & 70 & 30 \\
Context length & 218 & 58 & 3,369 & 1,167 & 352 & 4,278 & 1,410 & 1,397 \\
\midrule
\multicolumn{9}{c}{\textbf{OGWiC}} \\
\midrule
Instances & 6,833 & 10,833 & 8,279 & 5,910 & 4,504 & 8,029 & 4,821 & 5,457 \\
\midrule
\multicolumn{9}{c}{\textbf{DisWiC}} \\
\midrule
Instances & 11,740 & 20,461 & 13,690 & 10,831 & 6,041 & 12,698 & 9,339 & 9,117 \\
\bottomrule
\end{tabular}
\caption{Training set statistics of both tasks (OGWiC and DisWiC) per language (ISO codes in column headings) and on average (AVG, rounded to the nearest integer). The set of \emph{available} contexts and lemmas is identical in both tasks (top part), but the use of possible combinations differs in the two tasks, yielding varying amounts of training instances across tasks (bottom part). Unique contexts is the amount of unique contexts, unique lemmas the amount of unique words in consideration and context length is the average number of words per context (rounded to the nearest integer).}
\label{tab:dataset}
\end{table*}

\begin{table*}[hbtp]
\small
\centering
\begin{tabular}{lcccccccc}
\toprule
 & AVG & ZH & DE & EN & NO & RU & ES & SV \\
 \midrule
\multicolumn{9}{c}{\textbf{OGWiC} (Krippendorff's $\alpha$)} \\
\midrule
Baseline & 0.123 & 0.059 & 0.274 & 0.102 & 0.124 & 0.112 & 0.175 & 0.018 \\
XLM-R & 0.174 & 0.068 & 0.185 & 0.280 & 0.025 & 0.192 & 0.375 & 0.091 \\
XLM-R + Adapter & \textbf{0.340} & \textbf{0.187} & \textbf{0.396} & \textbf{0.394} & \textbf{0.283} & \textbf{0.341} & \textbf{0.435} & \textbf{0.347} \\
XLM-R + Ensemble & 0.242 & -0.052 & 0.199 & 0.347 & 0.217 & 0.316 & 0.330 & 0.337 \\
Top Submission & \underline{0.656} & \underline{0.424} & \underline{0.723} & \underline{0.723} & \underline{0.668} & \underline{0.623} & \underline{0.748}   & \underline{0.675} \\
\midrule
\multicolumn{9}{c}{\textbf{DisWiC} (Spearman's $\rho$)} \\
\midrule
Baseline & 0.118 & 0.387 & 0.093 & 0.064 & 0.076 & 0.049 & 0.077 & 0.081 \\
XLM-R & 0.083 & 0.398 & 0.067 & 0.016 & -0.118 & 0.045 & 0.052 & 0.119 \\
XLM-R + Adapter & 0.146 & 0.402 & 0.127 & \underline{\textbf{0.092}} & 0.113 & \textbf{0.091} & \textbf{0.103} & 0.097 \\
XLM-R + Ensemble & \textbf{0.170} & \underline{\textbf{0.433}} & \textbf{0.167} & 0.056 & \textbf{0.178} & 0.076 & 0.088 & \textbf{0.194} \\
Top Submission & \underline{0.226}  & 0.301  & \underline{0.204} & 0.078  & \underline{0.286} & \underline{0.175} & \underline{0.187}  & \underline{0.350}  \\
\bottomrule
\end{tabular}
\caption{Results on the test sets of both subtasks (OGWiC and DisWiC, evaluation metric in parentheses) per language (ISO codes in column headings) and on average (AVG). We compare our methods against the baselines provided by the task organizers (cf. section~\ref{ssec:baselines} and the best performing system (Deep Change) at the time of evaluation of the competition (indicated by ``Top Submission'' in the table). Best scores of our methods in \textbf{bold} and best overall \underline{underlined}.}
\label{tab:results}
\end{table*}

\section{Results}
In Table~\ref{tab:results}, we compare our three models (XLM-R, XML-R + Adapter, XML-R + Ensemble) to each other, to the baselines provided by the task organizers, and to the best performing submission in the shared task. Since the shared task is still open for participation, post-evaluation results are subject to change. Therefore, we compare against the official evaluation results from within the competition and report corresponding scores for the best submission. By average scores, our XLM-R + Adapter method would have ranked 5th in the OGWiC task and the XLM-R + Ensemble method 3rd in DisWiC.

In the OGWiC task, XLM-R + Adapter consistently performs best across all languages among our methods, but falls short in comparison to the best submission. On average, also the simple XLM-R method performs better than the baseline.

In the DisWiC task, best performance among our models varies between XLM-R + Adapter and XLM-R + Ensemble. While XLM-R + Ensemble outperforms the best submission on Chinese language and XLM-R + Adapter performs better than the best submission on English, scores of the best submission are highest on the remaining five languages and on average.
In comparison to the Linear Regression baseline as provided by the organizers, the addition of dropout in XLM-R seems to be harmful rather than helpful.

\section{Discussion}
Expectably, our methods with enriched features and more complex classifiers/regressor (XLM-R + Adapter and XLM-R) outperform our baseline of a simple classification/regression head directly on top of the concatenation of contextual embeddings (XLM-R). This behavior is consistent across languages, except for Chinese, where the XLM-R + Ensemble performs worst among all methods (including the CoMeDi baseline) in the OGWiC task. Generally, the subset of Chinese instances reveals interesting patterns. Despite that Chinese has the highest number of training instances in both tasks, performance is almost opposite between the two tasks: Chinese has the lowest score among almost all methods in OGWiC (and in particular the lowest score in the best submission), whereas it has the highest score among almost all methods in DisWiC (second-highest in best submission).
We hypothesize that this gap may be rooted in the set of available contexts, which is smallest for Chinese, despite Chinese having the highest amount of training instances in both tasks. That means, several contexts must appear in multiple instance whereas for example the Russian instances could be constructed almost exclusively from unique contexts (each instance is a pair of two contexts, i.e., 12698*2 = 25396 unique contexts would be required for every context to appear only once, whereas 24848 unique contexts are available). Since our methods build on contextual embeddings, for contexts that appear a lot of times, they might learn to rely on patterns in the corresponding contextual embeddings that are determined by context only and try to use these as shortcuts. This behavior might work in DisWiC, if the disagreement of annotators is governed by context rather than the lemma, but fail in the prediction of the relatedness of the actual lemma. However, that is only one potential explanation, while other components in the pipeline of our methods or differences in the task/data configuration may offer equally valid explanations. We also do not know details about the best performing submission and hence cannot judge whether that explanation would hold for it.

In the initial submission, we related the performance of individual methods to properties of the data for different languages, such as duplicated contexts. However, we noticed a mistake in the definition/calculation of duplicated contexts and that these conclusions were drawn erroneously. Therefore, we dropped this part of the discussion in the final submission.

\section{Conclusion}

In this shared task paper, we introduced multiple methods that incorporate extensions of contextual embeddings by pairwise comparison, such as element-wise difference and similarity measures, and additonal transformations of these embeddings by Adapter blocks to task-specific representations. We use the contextual embeddings (and their extensions)  with classifiers and regressors of varying complexity.

While the performance of our methods falls short in comparison to the best submission in the OGWiC task, it is competitive in terms of official evaluation results in the DisWiC task.

We are curiously looking forward to the descriptions of the other systems and plan to investigate potential options to combine approaches and ideas to advance future research on disagreement modeling in multilingual and multi-contextual settings.

\section*{Limitations}

This study focuses exclusively on WiC tasks involving seven specific languages, leaving the generalization of the models to other languages outside the scope of this shared task uncertain. Additionally, our approach is limited to the methods described in this work. Future research could explore the performance of these models across a wider range of languages and investigate the impact of alternative fine-tuning strategies on their overall effectiveness.

\section*{Acknowledgments}
Part of the research that led to this submission has been supported with funding by Hessian.AI. Any opinions, findings, conclusions, or recommendations in this material are those of the authors and do not necessarily reflect the views of Hessian.AI.

\bibliography{coling_latex}
\end{document}